%% file: main.tex
\documentclass[10pt,twocolumn,letterpaper]{article}

\usepackage{cvpr}              % To produce the CAMERA-READY version
\usepackage[accsupp]{axessibility}  % Improves PDF readability for those with disabilities.
\usepackage{multirow}
\usepackage{graphicx}  % Include this package in the preamble

\input{preamble}

\usepackage{xcolor} % for colored text

\definecolor{cvprblue}{rgb}{0.21,0.49,0.74}

\def\paperID{PHAROS-AIF-MIH-29} % *** Enter the Paper ID here
\def\confName{CVPR}
\def\confYear{2026}

\title{
Phase-map synthesis from magnitude-only MR images using conditional score-based diffusion models with application in training of accelerated MRI reconstruction models
}

\author{M.~Berk~Sahin$^*$
	\qquad~Dilek~Yalcinkaya$^*$
	\qquad~Abolfazl Hashemi
    \qquad~Behzad~Sharif $ ^\dagger$
    \\
	{Purdue University}
	\\
	\small{\texttt{\{sahinm, dyalcink, abolfazl, bsharif\}@purdue.edu}}
}

\begin{document}

\makeatletter
\twocolumn[{%
\renewcommand\twocolumn[1][]{#1}

\begin{center}
\begin{minipage}{0.92\textwidth}
\vspace{-1em}
{\color{red}\bfseries\large
Accepted for publication in Proceedings of IEEE/CVF Conference on Computer Vision and Pattern Recognition (CVPR) Workshop. The final version will be available on IEEE Xplore shortly after CVPR 2026.
}

\vspace{0.7em}

{\color{red}\large
Cite as: Sahin MB, et al. ``Phase-map synthesis from magnitude-only MR images using conditional score-based diffusion models with application in training of accelerated MRI reconstruction models'', in Proceedings of the Computer Vision and Pattern Recognition Conference, 2026.
}

\end{minipage}
\end{center}

\vspace{-3em}

\maketitle
}]
\makeatother

\begingroup
\renewcommand\thefootnote{\fnsymbol{footnote}}
\renewcommand\footnoterule{}
\makeatletter
\long\def\@makefntext#1{\noindent{\footnotesize #1}}
\makeatother
\footnotetext[1]{$^*$Equal contribution.}
\footnotetext[2]{$^\dagger$Corresponding author.}
\endgroup

\input{sec/0_abstract}

\input{sec/1_intro}
{
    \small
    \bibliographystyle{ieeenat_fullname}
    \bibliography{main}
}


\end{document}

%% file: preamble.tex
\usepackage{dblfloatfix}
\usepackage{bm}
\usepackage{amsfonts}
\usepackage{amsmath}
\usepackage{amssymb}
\usepackage{cite}
\usepackage{multirow}
\usepackage{subcaption}
\usepackage{floatrow}
\usepackage[pagebackref,hidelinks]{hyperref} % Include the hyperref package
\usepackage{array, caption, floatrow, tabularx, makecell, booktabs}% 
\usepackage{caption}
\usepackage{cuted}

\newcommand{\x}{{\bm x}}
\newcommand{\y}{{\bm y}}
\newcommand{\w}{{\bm w}}
\newcommand{\f}{{\bm f}}

\newcommand{\I}{{\bm I}}
\newcommand{\E}{\mathbb{E}}
\newcommand{\R}{\mathbb{R}}
\newcommand{\defeq}{\mathrel{\mathop:}=}

%% file: sec/0_abstract.tex
\begin{abstract}

%To alleviate the difficulty of creating large-scale annotated datasets for abnormality classification in medical images that is a labor-intensive and costly endeavor, the scarcity of annotated data, weak supervision from readily available resources, such as radiology reports, can provide an effective alternative for training abnormality classification models. However, most of the current wor
\vspace{-1em}
Accelerated magnetic resonance imaging (MRI) enabled by the training of deep learning (DL)-based image reconstruction models requires large and diverse raw k-space (Fourier domain) datasets. In most clinical MRI applications, due to storage and patient privacy concerns, raw k-space data is discarded and magnitude-only images are the only component saved. Consequently, a large portion of the DL-based MRI reconstruction literature has either relied on small training datasets or has used one of the few available open-source k-space datasets. At the same time, the growing number of anonymized magnitude-only image registries/databases motivates the development of techniques that can use them as training datasets for generalizable DL-based reconstruction models. Here we propose to address this challenge by employing a generative approach based on conditional score-based diffusion models (SBDMs): given a magnitude-only MR image, it synthesizes a phase map (in the image domain) that realistically corresponds to the magnitude-only image. We evaluate its generative capabilities in a downstream DL-based reconstruction task whereby a large k-space dataset is generated by combining the SBDM-synthesized phase-maps and the corresponding magnitude-only images, and this k-space dataset is then used to train a DL model for accelerated MRI reconstruction. We compare the performance of the resulting DL model versus those trained according to (a) a naïve approach that uses smooth phase, (b) a k-space training dataset generated using synthesized phase maps derived from a generative adversarial network, and (c) the ground truth k-space data. Our results suggest that the DL model trained from SBDM-synthesized k-space data outperforms the other approaches in terms of quantitative metrics as well as qualitatively observed reconstruction fidelity, i.e., whether the reconstructed images include erroneous or hallucinated features that could adversely impact diagnostic accuracy.

% I will make edits later -Berk

\end{abstract}

%% file: sec/1_intro.tex
\begin{figure*}[t]
\centering
\includegraphics[width=7in]{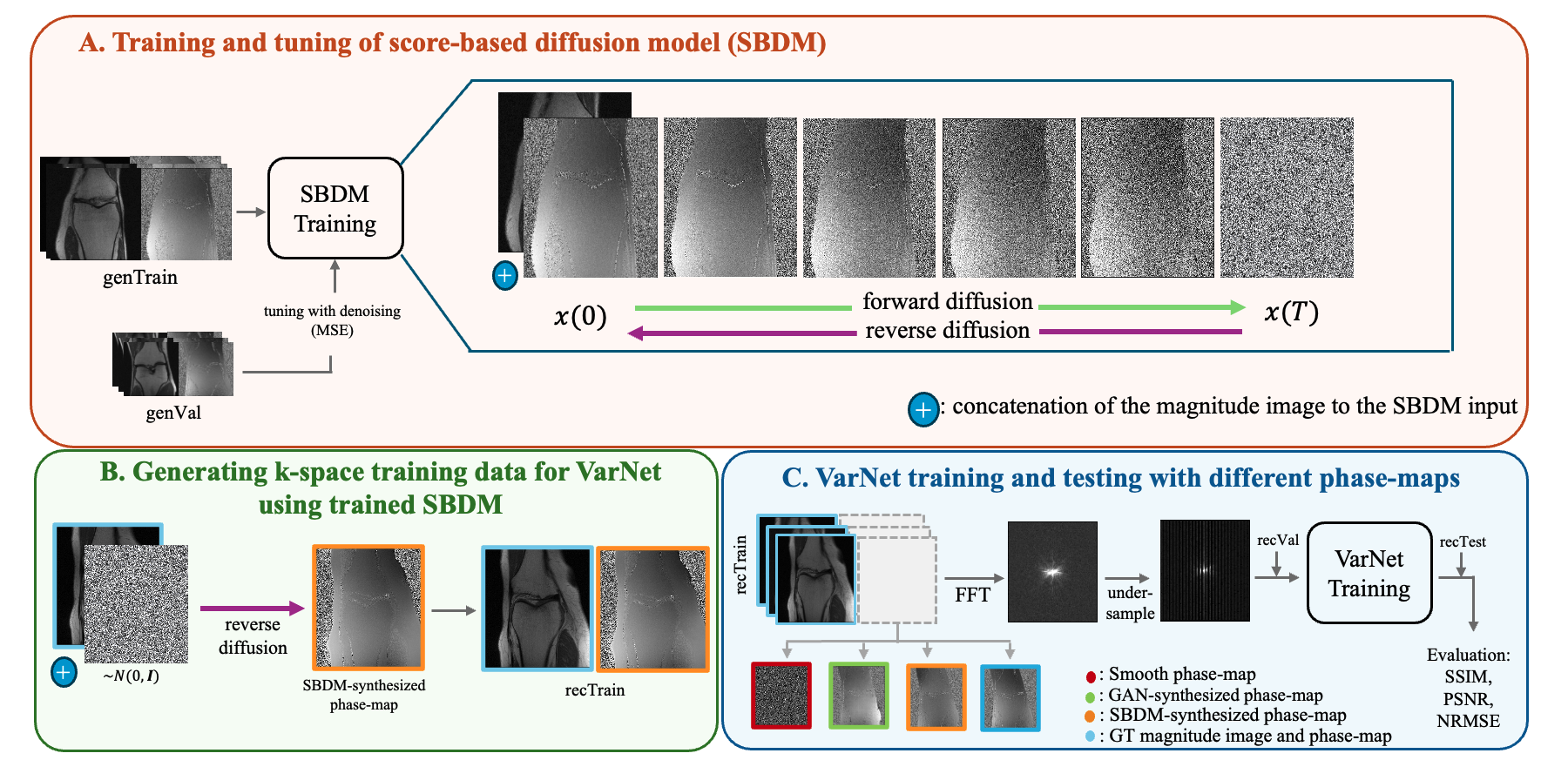}
\caption{Pipeline for the proposed phase-map synthesis approach. Following the training of the SBDM in (A), large k-space datasets can be synthesized (recTrain) by generating realistic phase-maps that correspond to the magnitude-only MR images as shown in (B). In (C), the generative performance of the proposed method is compared vs. alternative methods by training a VarNet model for accelerated MRI reconstruction using real-world testing data (recTest).}
\label{fig_method}
\end{figure*}
\vspace{-1em}
\section{Introduction} 
Deep learning (DL)-based techniques have been shown to be effective in solving the inverse problems associated with image reconstruction in accelerated magnetic resonance imaging (MRI). Real-world deployment of such techniques, however, has been limited which can be partially attributed to issues involving generalizability and performance degradation in the face of ``dataset shifts'' which can be, in part, attributed to limitations of the training dataset. A major obstacle in curating large and diverse multi-vendor multi-site datasets for MRI reconstruction is that nearly all clinical sites discard the raw k-space data and only retain the magnitude-only images in most imaging protocols, typically stored in the standard DICOM format. Although applying a simple forward Fourier transform to a magnitude-only image results in a ``valid'' k-space data, discarding the phase information in the raw k-space during the training process ignores the features that are contained in the phase map (which is inherently nonzero in MRI) and may result in subpar reconstruction performance.

It is known that using synthetic data improves generalization performance in the context of DL-based accelerated MRI reconstruction \citep{deveshwar2023synthesizing, wang2023one}, MRI segmentation \citep{al2023usability}, and X-ray \citep{gao2023synthetic}. This motivates the need to develop generative models to synthesize complex raw k-space training sets by leveraging magnitude-only MRI  databases for improving image reconstruction. Generative adversarial networks \citep{goodfellow2020generative} is one of the state-of-the-art approaches in generative tasks; however, they are known to suffer from training instability and yield biased results \citep{bau2019seeing, maluleke2022studying}. An alternative to GANs is the recently emerging diffusion models with their impressive performance \citep{rombach2022high, song2021score,chung2022score,ho2020denoising,dhariwal2021diffusion}. During conditional image generation, the diffusion process is guided by a reference image, which was studied for medical image synthesis \citep{peng2022generating,jiang2023cola,brain_gen}. Still, the feasibility of the diffusion models for synthesizing MRI raw k-space from magnitude-only images is currently unexplored.

We present a novel score-based diffusion model (SBDM) method for generating realistic MRI phase-maps from magnitude-only images. In our approach, conditioning on the magnitude image is enabled by its concatenation to the SBDM input.  We evaluate the synthesized phase-maps by using them to train a downstream DL-based image reconstruction task. Our results suggest that the proposed approach outperforms the state-of-the-art GAN method across various acceleration regimes in terms of reconstruction quality metrics and reconstruction fidelity. Furthermore, the DL model trained with the SBDM-synthesized phase-maps shows similar performance vs. the one trained with ground-truth phase-maps. To the best of our knowledge, this is the first work to explore applying SBDMs for phase-maps synthesis from magnitude-only MR images.

\begin{figure*}[t]
\centering
\includegraphics[width=7in]{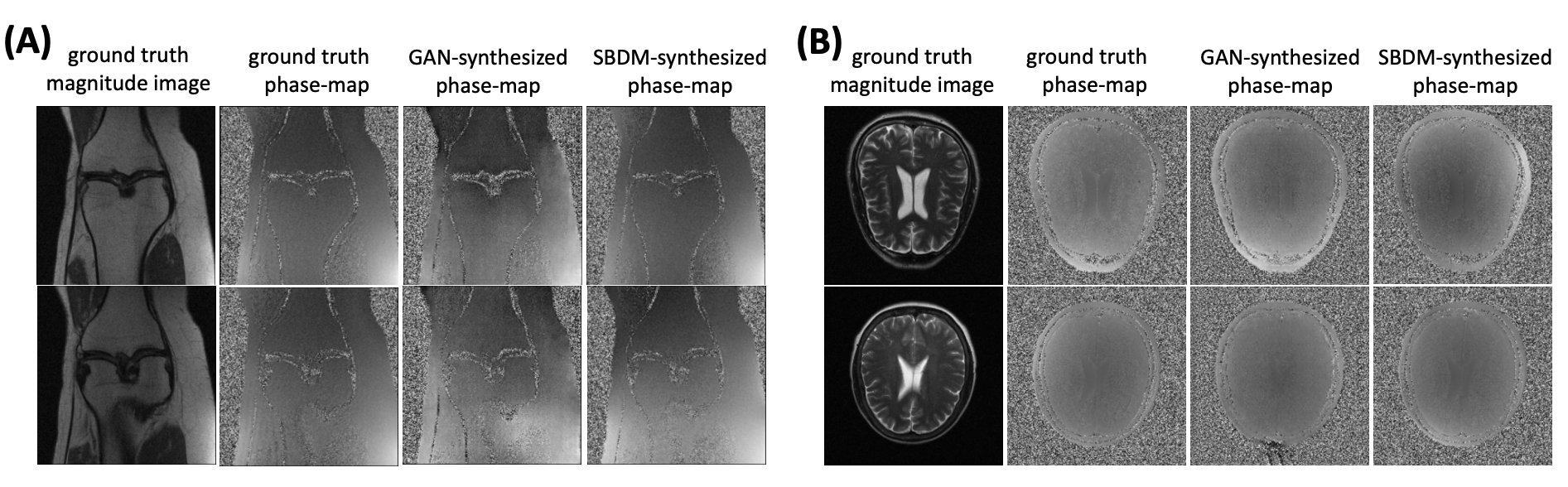}
\caption{Examples of GAN- and SBDM-synthesized phase-maps from the magnitude-only images for (A) knee and (B) brain datasets. The phase maps generated by GAN and SBDM are realistic (not identical) when compared to the GT and show slightly different characteristics.}
\label{fig_phaseEx}
\end{figure*}

\begin{table*}[t]
  \floatsetup{floatrowsep=qquad, captionskip=4pt}
  \begin{floatrow}[2]
    \ttabbox%
    {%
    \footnotesize
    \begin{tabularx}{0.48\textwidth}{l >{\centering\arraybackslash}X >{\centering\arraybackslash}X}
      \toprule
      & Knee & Brain \\
      \cmidrule(lr){2-2}\cmidrule(lr){3-3}
      genTrain & 330/24,854 & 1928/30,494 \\
      genVal   & 30/2,297   & 149/2,372 \\
      genTest  & 196/7,591  & 1024/16,202 \\
      \bottomrule
    \end{tabularx}
    }%
    {\caption{Initial dataset split for SBDM (patient/image).}\label{table:gen_split}}
    \hfill
    \ttabbox%
    {%
    \footnotesize
    \begin{tabularx}{0.48\textwidth}{l >{\centering\arraybackslash}X >{\centering\arraybackslash}X}
      \toprule
      & Knee & Brain \\
      \cmidrule(lr){2-2}\cmidrule(lr){3-3}
      recTrain & 126/4,952 & 729/11,526 \\
      recVal   & 30/1,435  & 99/1,562 \\
      recTest  & 40/1,204  & 196/3,114 \\
      \bottomrule
    \end{tabularx}
    }%
    {\caption{Second split on genTest to train VarNet (patient/image).}\label{table:rec_split}}
  \end{floatrow}
\end{table*}

\section{Methods}

\subsection{Datasets}\label{section: dataset}

Knee and brain MRI data from the publicly available\footnote{\href{https://fastmri.med.nyu.edu/}{https://fastmri.med.nyu.edu/}} fastMRI dataset~\citep{zbontar2018fastmri, knoll2020fastmri} were used for training of the score-based diffusion model (SBDM) and its evaluation with a DL-based reconstruction framework. Both of these datasets included the ground truth k-space data. The k-space data for the knee dataset was available in a single-channel format while multi-channel k-space from the brain dataset was compressed into a single channel based on prior work~\citep{tygert2020simulating}. The knee and brain datasets were split into three groups for SBDM: training (genTrain), validation (genVal), and test (genTest) as described in Table \ref{table:gen_split}. Next, genTest was split into three subgroups once more to establish the dataset split for the reconstruction network VarNet, i.e., genTest = recTrain + recVal + recTest (described in Table \ref{table:rec_split}). Images in both the brain and knee datasets were cropped to the center of the raw k-space and resized to $256\times256$ in the image domain.

\subsection{Overview of the proposed method}
Fig. \ref{fig_method} summarizes our proposed pipeline for phase-map synthesis. In (A), an SBDM is trained by concatenating the ground-truth (GT) magnitude images and phase-maps using genTrain split. Then, hyperparameter tuning is performed by evaluating the denoising performance \citep{song2021score} on genVal. In (B), the trained SBDM is used for synthesizing the k-space data from magnitude-only images. In (C), a DL-based reconstruction framework, a VarNet model \citep{varnet}, was trained using the synthesized k-space data to evaluate the ability of the proposed approach to generate realistic k-space data through phase-map synthesis. The same procedure in (C) was repeated using alternative approaches to provide a comparison: smooth, GAN, and GT phase-maps, each providing a different VarNet model. In the final step, these VarNets are tested with real-world k-space data (recTest) and the reconstruction image quality is reported using structural similarity index measure (SSIM) \citep{ssim}, peak signal-to-noise ratio (PSNR), and normalized root mean square error (NRMSE) metrics.

\subsection{Phase-map synthesis with SBDMs}

A continuous diffusion process $\{\x(t)\}^T_{t=0}$ can be constructed with $\x(t)\in\R^n$, where $t\in [0,T]$ is the time index of the process and $n$ denotes the image dimension. We set $\x(0)\sim p_{0}$ and $\x(T)\sim p_T$, where $p_{0}$ is the data distribution with i.i.d. samples and $p_T$ has a tractable form (i.e., isotropic Gaussian distribution) for efficient sample generation. We define $p_{0}\defeq p(\x|\y)$ as the posterior distribution of phase-maps, $\x\in\R^n$, conditioned on magnitude images, $\y\in\R^n$. Then, the stochastic process can be constructed as the solution of the following stochastic differential equation (SDE)
\begin{equation}\label{forward_sde}
    d\x = \f(\x, t)dt + g(t)d\w, 
\end{equation}
where $\f:\R^n\rightarrow\R^n$ is the drift term, $g:\R\rightarrow\R$ is the diffusion coefficient, $dt$ is the infinitesimal time step, and $\w$ is standard $n$-dimensional Wiener process. 
\par 
%There are two common way of choosing $\f$ and $g$ functions in the literature. %First, one may choose $\f(\x,t) = -\frac{1}{2}\beta(t)\x$, $g(t) = \sqrt{\beta(t)}$, where $0 < \beta(t) < 1$ is monotonically increasing function, and as $t\rightarrow\infty$, 
There are two common choices of $\f$ and $g$ functions in the literature. They can be chosen such that the magnitude of the signal decays to 0 and variance is preserved to a fixed constant as $t\rightarrow\infty$, which is called variance preserving (VP)-SDE~\citep{ho2020denoising} and it can be seen as the continuous version of DDPM \citep{song2021score}. 
%Alternatively, by choosing $\f(\x,t) = \bm{0}$, $g(t) = \sqrt{d[\sigma^2(t)]/dt}$, where $\sigma(t)>0$ is again a monotonically increasing function, one can construct a variance exploding (VE)-SDE. $\sigma(t)$ is generally chosen to be a geometric series~\cite{song2021score, kingma2021variational}. As the name suggests, contrary to VP-SDE, variance of the signal gets very large as $t\rightarrow\infty$. 
Alternatively, $\f$ and $g$ can be chosen in such a way that the variance of the signal gets very large as $t\rightarrow\infty$, which is called variance exploding (VE)-SDE \citep{song2021score}. Since we empirically observed that VE-SDE provides better sample qualities than VP-SDE, similarly observed in~\citep{chung2022score}, we used VE-SDE in our proposed method. 

\begin{figure*}[t]
    \centering
    \includegraphics[width=5in]{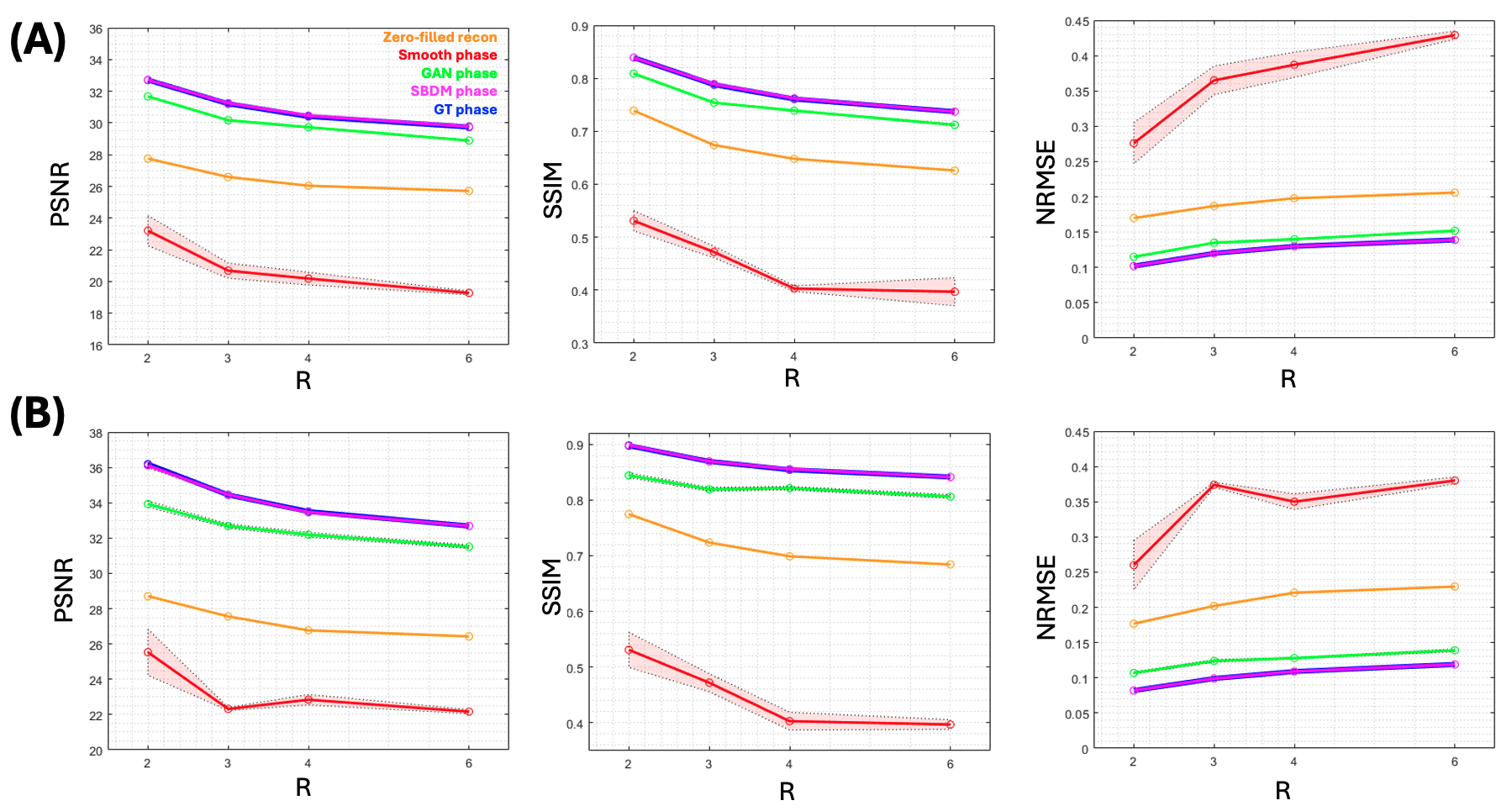}
    \caption{Cumulative VarNet reconstruction results with $n_\text{ACS}=26$ across a variety of acceleration factors ($\text{R}\in\{2,3,4,6\}$) for the (A) knee  and (B) brain datasets.}
    \label{fig:fig_cumul}
\end{figure*}

\begin{figure*}[t]
\centering
\includegraphics[width=5in]{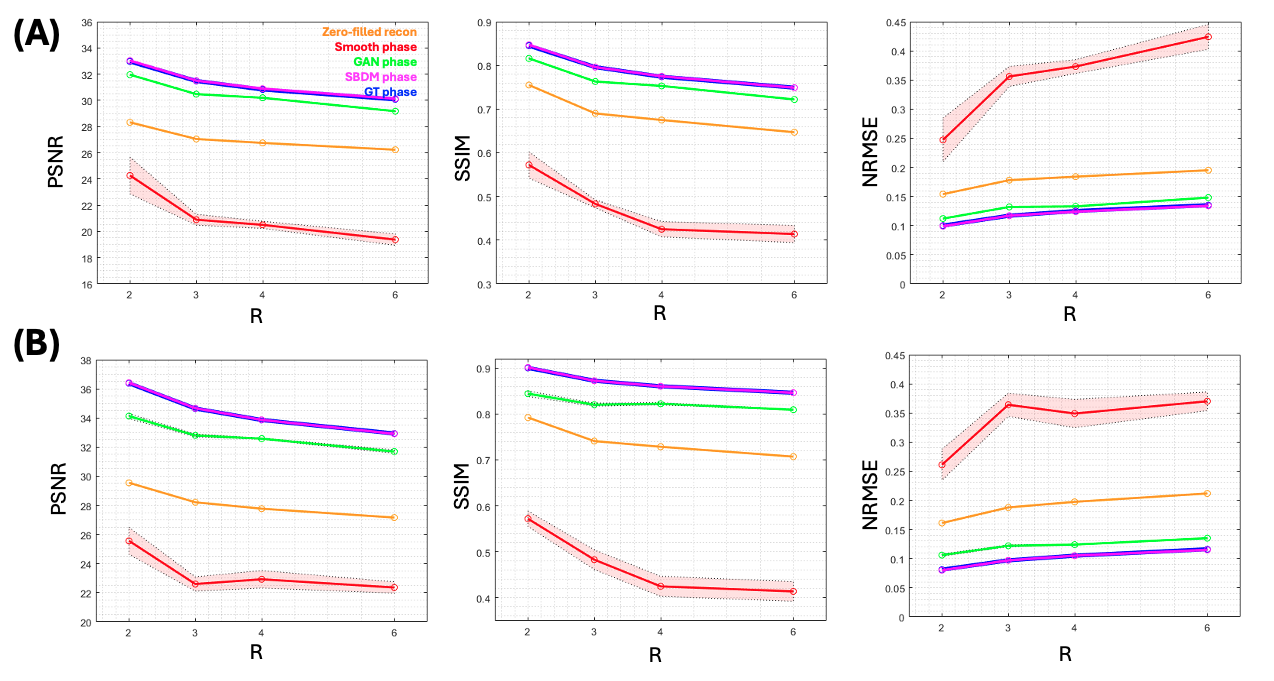}
\captionof{figure}{Cumulative VarNet reconstruction results with $n_\text{ACS}=31$ across a variety of acceleration factors ($\text{R}\in\{2,3,4,6\}$) for (A) knee and (B) brain datasets.}
\label{figure: cumul2}
\end{figure*}

\begin{figure*}[t]
\centering
\includegraphics[width=5in]{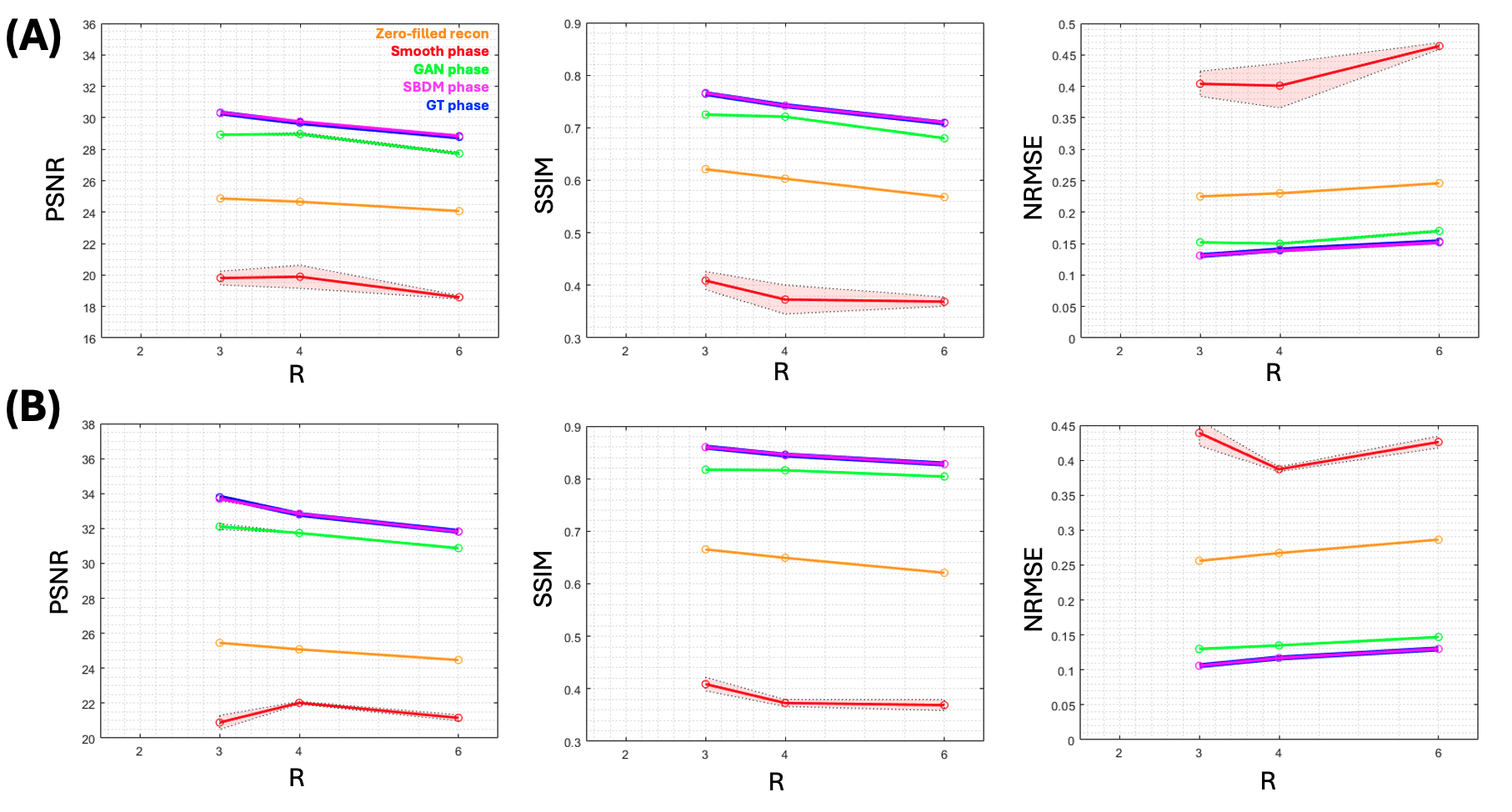}
\captionof{figure}{Cumulative VarNet reconstruction results with $n_\text{ACS}=16$ across a variety of acceleration factors ($\text{R}\in\{3,4,6\}$) for (A) knee  and (B) brain datasets.}
\label{figure: cumul3}
\end{figure*}

\par The reverse process of \eqref{forward_sde} can be constructed by another SDE~\citep{song2021score}: 
\begin{equation}\label{reverse_sde}
    d\x = [\f(\x, t) -g(t)^2\underbrace{\nabla_x\log p_t(\x|\y)}_{\text{score function}}]dt + g(t)d\Tilde{\w}, 
\end{equation} 
where $d\Tilde{\w}$ is the standard $n$-dimensional Wiener process in reverse direction. One can estimate the score function with a neural network conditioned on time indices and magnitude images, which is denoted by $s_{\theta}(\x(t), t, \y))\simeq \nabla_x\log p_t(\x|\y)$. We used sliced score matching \citep{vincent} to train our SBDM by minimizing the loss:

%$\nabla_{\x} \log p_{0t}(\x(t)|\x(0), \y)$ instead of $\nabla_{\x} \log p_{t}(\x|\y)$, where $p_{0t}(\x(t)|\x(0), \y)$ is pre-specified Gaussian noise distribution such that $p_{t}(\x|\y)\triangleq\int p_{0t}(\x(t)|\x(0), \y)p(\x|\y)$ for $t\in\R$. Formally, the loss function of SBDM can be written as:
%\begin{equation}\label{formal_loss_function}
%\begin{split}
    %L(\theta) \triangleq \E_{t\sim U[0,1]}\biggl[\lambda(t)\E_{\x(0), \y\sim p_{data}}& \E_{x(t)\sim p(\x(t)|\x(0))}\biggl[ \\
    %&\biggl.\lVert s_{\theta}(\x(t), t, \y) - \nabla_{\x}\log p_{0t}(\x(t)|\x(0), \y)\rVert^2_2\biggr]\biggr],
%\end{split}
%\end{equation}where $\lambda(t)$ is a weighting function, which we choose $\lambda(t)=\sigma^2(t)$ as suggested in ~\cite{song2021score}. Since we have a Gaussian distribution, we know that $\nabla_{\x}\log p_{0t}(\x(t)|\x(0), \y) = (\x(t) - \x(0))/\sigma^2(t)$. Combining these, we train our network with the following loss:
\begin{equation}
\label{train_loss}
\begin{split}
L(\theta) =\;& \E_{t\sim U[\epsilon,1]} \\
&\E_{(\x(0),\y)\sim p_{\mathrm{data}}}
\E_{\x(t)\sim\mathcal{N}(\x(0),\,\sigma^2(t)\I)}
\Biggl[ \\
&\qquad
\left\lVert
\sigma(t)s_{\theta}\bigl(\x(t), t, \y\bigr)
-
\frac{\x(t)-\x(0)}{\sigma(t)}
\right\rVert_2^2
\Biggr].
\end{split}
\end{equation}
where $p_{data}\defeq p(\x, \y)$ is the joint distribution, $\sigma(t)$ is a positive weighting term, and $\epsilon=10^{-5}$ is for preventing numerical issues \citep{song2021score}. Intuitively, we train a time-conditional neural network to denoise noisy ground-truth phase-maps concatenated with magnitude images. After training is done, in the phase-map generation stage, we solve \eqref{reverse_sde} by replacing the score function with $s_{\theta}(\x(t), t, \y))$ for every $t$ to generate a phase-map corresponding to the given magnitude image.

For the proposed SBDM approach, we used the implementation of the time-dependent score function model \texttt{ncsnpp}~\citep{song2021score, chung2023solving}, which is based on U-Net \citep{ronneberger2015u} and its sub-blocks are adopted from residual blocks of BigGAN \citep{biggan}. Skip-connections of the residual blocks are scaled by $1/\sqrt{2}$ as in~\citep{karras1, karras2}. Batch size of 12 and 32 was used for knee and brain datasets, respectively, $T=1000$, noise variance schedule of $\sigma(t)=\sigma_{min}(\sigma_{max}/\sigma_{min})^t$, where $\sigma_{min}=0.01$ and $\sigma_{max}=378$ as advised in \citep{improved_techs}. We use Adam optimizer \citep{adam} with linear warm-up schedule, reaching learning rate of ${10}^{-4}$ in ${5000}^{th}$ step. We compared our SBDM approach with the GAN described in \citep{deveshwar2023synthesizing}, which has a 16-layer U-Net generator  \citep{ronneberger2015u} with skip connections and $70\times 70$ PatchGAN discriminator \citep{isola2017image}. Implementation details for GAN can be found in \citep{isola2017image}. SBDM and GAN were trained for 60 and 100 epochs, respectively, and the best models were selected by MSE score and visual inspection of the samples generated on genVal set.

\subsection{Synthesized phase-map quality assessment}
We trained the VarNet as a downstream task to assess the phase-map quality by combining the magnitude image with the synthesized (smooth a.k.a. na\"{i}ve, GAN-synthesized, SBDM-synthesized), and real phase-maps. Smooth phase-maps are sampled from Gaussian distribution. Frechet Inception Distance (FID) \citep{fid, fid_implement} was also computed on recTrain for GAN- and SBDM-derived phase-maps.

VarNet was trained with the Adam optimizer \citep{adam}, learning rate of $3\times10^{-4}$, and batch size of 4. The main difference between the original VarNet implementation and ours was that we altered the VarNet architecture to accommodate for the single-coil input. Implementation details (e.g., number of cascades etc.) can be found in \citep{varnet}. Our code for the proposed pipeline can be found here\footnote{\href{https://github.com/BerkMSahin/phase-map-synthesis-with-SBDM}{https://github.com/BerkMSahin/phase-map-synthesis-with-SBDM}}.
\section{Results}
\subsection{Phase-map synthesis results}
Fig. \ref{fig_phaseEx} shows two slices of one subject from (A) brain and (B) knee datasets. The results for both generative models (GAN and SBDM) show several expected features, including tissue phase contrast, low spatial-frequency components, and noisy background. In the bottom row of (B), some superfluous sharp patterns appear in the GAN-derived phase-map that are not present in GT or the SBDM-synthesized phase-maps. For the knee dataset, FID score on the synthesized phase-maps of recTrain dataset for GAN and our proposed SBDM approach was 40.65 and 5.46, respectively. For the brain dataset, this was 32.08 for GAN and 3.46 for SBDM.

\begin{figure*}[t]
\centering
\includegraphics[width=7in]{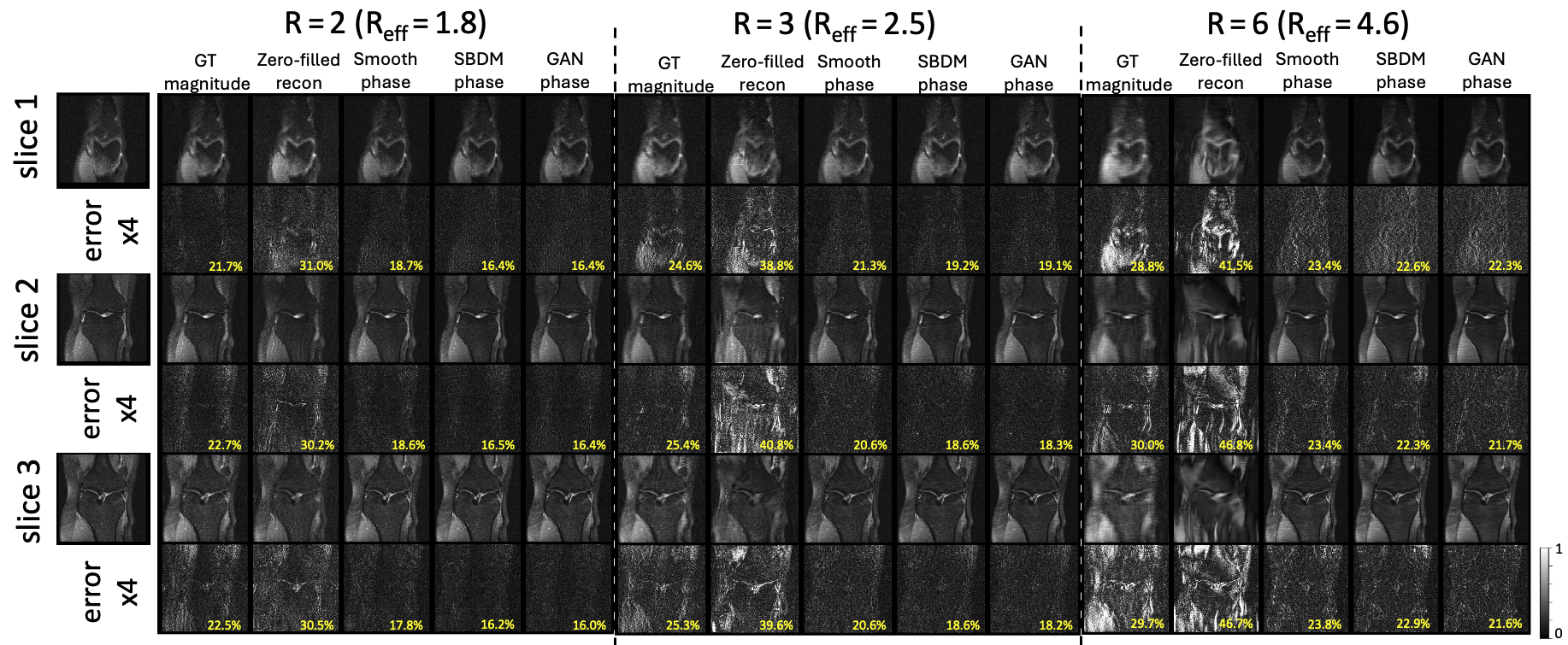}
\caption{VarNet reconstruction results for a subject from the knee dataset are shown at three acceleration factors. The NRMSE corresponding to each reconstruction-error map is shown in yellow.}
\label{figure: repCase1}
\end{figure*}

\begin{figure*}[t]
\centering
\includegraphics[width=7in]{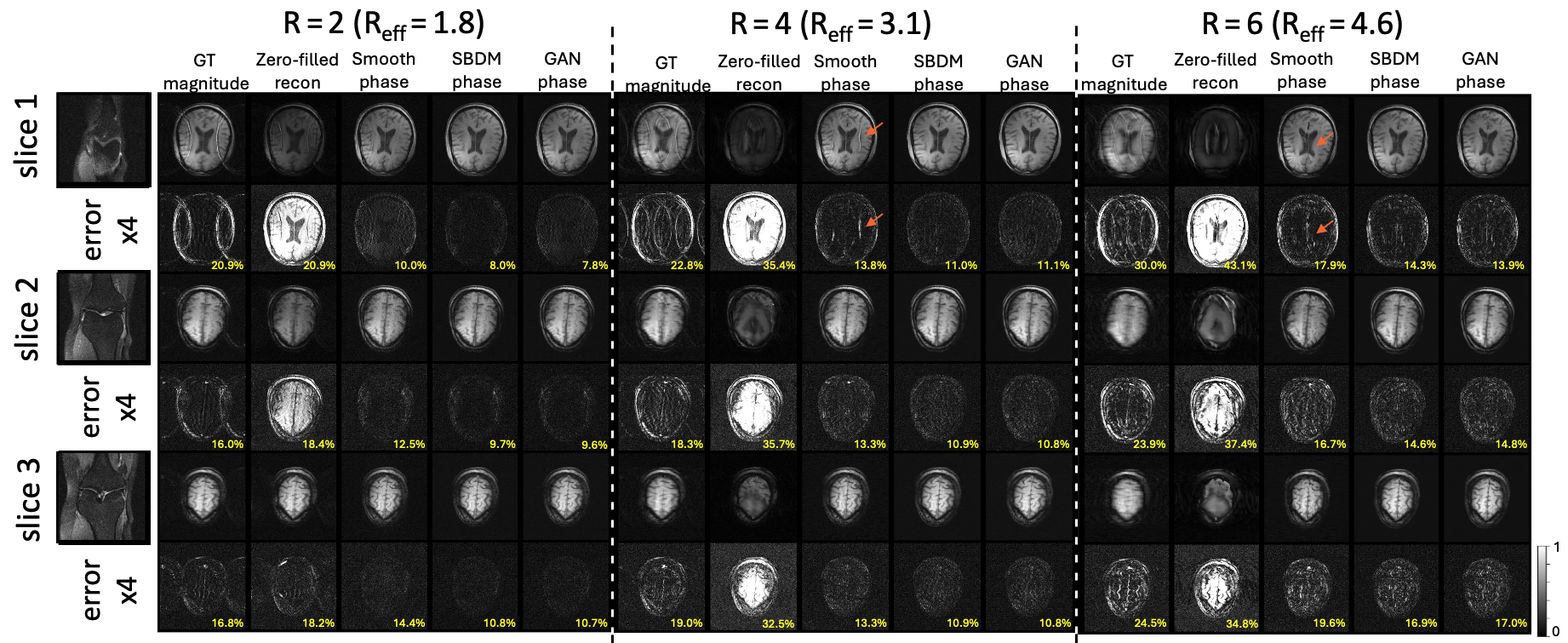}
\caption{VarNet reconstruction results for a subject from the brain dataset are shown at three acceleration factors. The NRMSE corresponding to each reconstruction-error map is shown in yellow. Orange arrows point to the hallucinated features.}
\label{figure: repCase2}
\end{figure*}

\subsection{Reconstruction cumulative results}

When training VarNet, we experiment with a combination of different acceleration factors ($\text{R}$) and the size of the fully sampled central k-space region (auto-calibration region; $n_\text{ACS}$). Fig. \ref{fig:fig_cumul} summarizes the cumulative results for the (A) knee and (B) brain datasets wherein the average performance of three random seeds is reported. Additional cumulative results with $n_\text{ACS}\in\{16,31\}$ is reported in Figs.~\ref{figure: cumul2} and~\ref{figure: cumul3}. VarNet trained with SBDM-synthesized phase-maps performs nearly the same as the model trained with actual k-space. ``Smooth phase'' refers to the baseline approach where the magnitude image is combined with Gaussian phase map to obtain k-space. We also provide the zero-filled reconstruction (IFFT) results as the naïve reconstruction approach.

\begin{figure*}[t]
\centering
\includegraphics[width=7in]{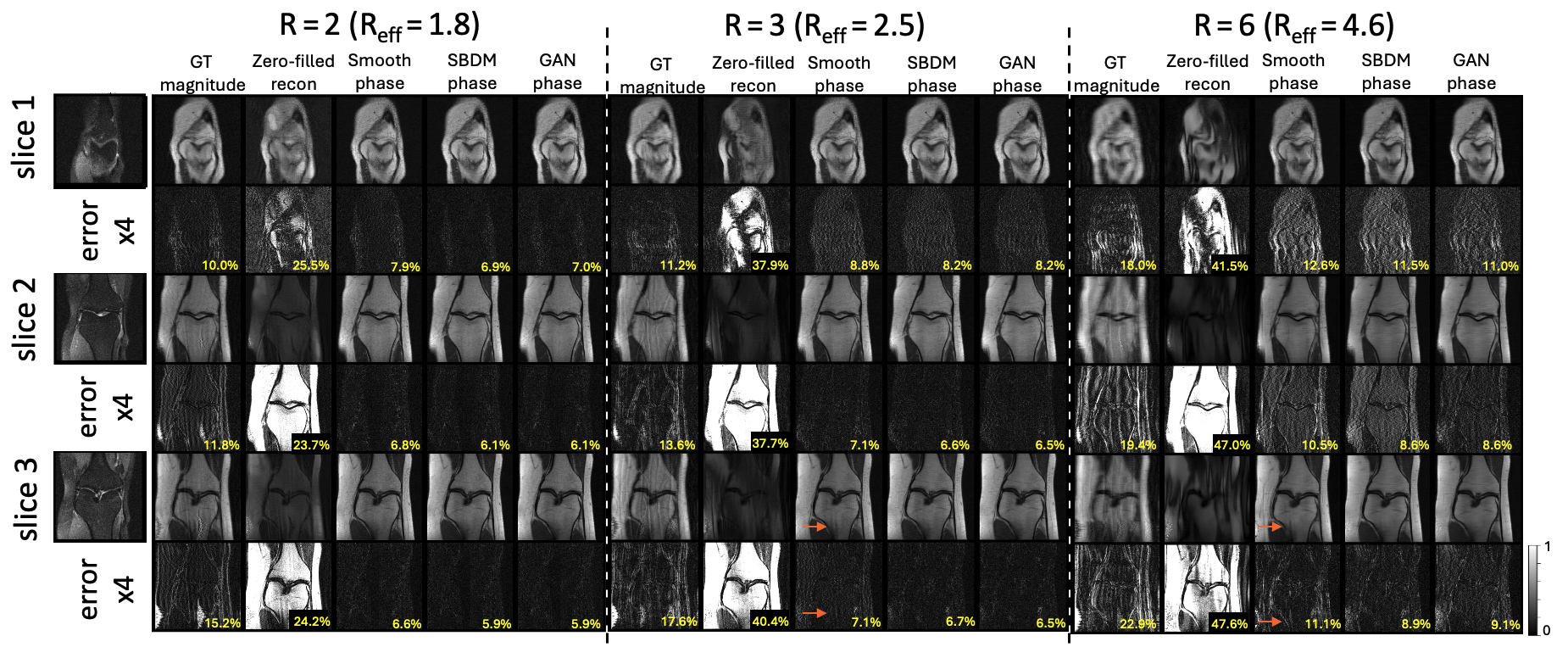}
\caption{VarNet reconstruction results for a subject from the knee dataset is shown at three acceleration factors. The NRMSE corresponding to each reconstruction-error map is shown in yellow. Orange arrows point to the hallucinated features.}
\label{figure: repCase3}
\end{figure*}

\subsection{Reconstruction representative cases}
Figs.~\ref{figure: repCase1} and~\ref{figure: repCase2} show the representative test cases from the knee and brain datasets, respectively, for three slices and three different ``effective acceleration'' factors (R$_\text{eff}$), which is computed by taking into account the fully sampled calibration region. Although the model trained with GAN-synthesized phase-maps has comparable performance in terms of NRMSE, its reconstruction quality is worse than our SBDM approach (visual inspection, arrows, error maps), and this gap in reconstruction quality widens in higher effective accelerations. In Fig. \ref{figure: repCase2}, it can be seen that the VarNet model trained with GAN-synthesized phase-maps generates ``hallucinations'' on brain structures, highlighted by orange arrows. Additional representative cases are presented in Fig.~\ref{figure: repCase3}. 

\section{Discussion and Conclusion}

We proposed to leverage the emerging developments in the field of generative diffusion models, specifically score-based diffusion models, for retrospective phase-map synthesis from magnitude-only knee or brain magnetic resonance (MR) images. Our proposed technique has the potential to create large k-space (Fourier domain) MRI datasets using widely available registries or databases with magnitude-only MR images to improve deep learning-based accelerated MRI reconstruction. 
For low effective acceleration (R$_{\text{eff}}$) levels, VarNet trained with smooth phase-maps (naïve approach) has comparable reconstruction performance (e.g., see the trend observed in Fig. \ref{figure: repCase1}). However, as R$_{\text{eff}}$ increases, reconstruction quality degrades severely for the naïve approach whereas the reconstruction performance for VarNet models trained with phase-maps synthesized by GANs or SBDMs show a much more benign level of degradation. This implies that having a ``realistic" phase map is crucial to be able to utilize magnitude-only MR image databases efficiently for training highly accelerated reconstruction models. Moreover, there was a notable difference between the FID scores of GAN and SBDM, suggesting the potential superiority of the SBDM approach in this context. This is consistent with the qualitative comparison shown in Fig. \ref{fig_phaseEx} where some erroneous patterns are observed at the edges of the noisy background of the GAN-derived phase-maps which are not present in the ground truth and SBDM-derived phase-maps. Such differences in the FID scores between GANs and diffusion models were also observed in~\citep{brain_tumor_seg, fid_explanation2, brain_gen}.

\subsubsection*{Acknowledgments} This work was partially supported by National Institutes of Health grant NIH/NHLBI R01-HL153430 (PI: Sharif).

%% file: main.bib
@String(CVPR= {IEEE Conf. Comput. Vis. Pattern Recog.})

@String(ICLR = {Int. Conf. Learn. Represent.})

@String(CVPR  = {CVPR})

@String(ICLR  = {ICLR})

@article{ho2020denoising,
  title={Denoising diffusion probabilistic models},
  author={Ho, Jonathan and Jain, Ajay and Abbeel, Pieter},
  journal={Advances in neural information processing systems},
  volume={33},
  pages={6840--6851},
  year={2020}
}

@inproceedings{song2021score,
  title = {Score-Based Generative Modeling through Stochastic Differential Equations},
  author = {Yang Song and Jascha Sohl-Dickstein and Diederik P. Kingma and Abhishek Kumar and Stefano Ermon and Ben Poole},
  cites = {0},
  citedby = {0},
  booktitle = {9th International Conference on Learning Representations, ICLR 2021.}, 
  year={2021}
}

@article{chung2022score,
  title={Score-based diffusion models for accelerated MRI},
  author={Chung, Hyungjin and Ye, Jong Chul},
  journal={Medical image analysis},
  volume={80},
  pages={102479},
  year={2022},
  publisher={Elsevier}
}

@ARTICLE{vincent,
  author={Vincent, Pascal},
  journal={Neural Computation}, 
  title={A Connection Between Score Matching and Denoising Autoencoders}, 
  year={2011},
  volume={23},
  number={7},
  pages={1661-1674}}

@article{zbontar2018fastmri,
  title={fastMRI: An open dataset and benchmarks for accelerated MRI},
  author={Zbontar, Jure and Knoll, Florian and Sriram, Anuroop and Murrell, et al.},
  journal={arXiv preprint arXiv:1811.08839},
  year={2018}
}

@article{tygert2020simulating,
  title={Simulating single-coil MRI from the responses of multiple coils},
  author={Tygert, Mark and Zbontar, Jure},
  journal={Communications in Applied Mathematics and Computational Science},
  volume={15},
  number={2},
  pages={115--127},
  year={2020},
  publisher={Mathematical Sciences Publishers}
}

@inproceedings{ronneberger2015u,
  title={U-net: Convolutional networks for biomedical image segmentation},
  author={Ronneberger, Olaf and Fischer, Philipp and Brox, Thomas},
  booktitle={Medical Image Computing and Computer-Assisted Intervention--MICCAI 2015: 18th International Conference, Munich, Germany, October 5-9, 2015, Proceedings, Part III 18},
  pages={234--241},
  year={2015},
  organization={Springer}
}

@inproceedings{chung2023solving,
  title={Solving 3d inverse problems using pre-trained 2d diffusion models},
  author={Chung, Hyungjin and Ryu, Dohoon and McCann, Michael T and Klasky, Marc L and Ye, Jong Chul},
  booktitle={Proceedings of the IEEE/CVF conference on computer vision and pattern recognition},
  pages={22542--22551},
  year={2023}
}

@inproceedings{biggan,
  author       = {Andrew Brock and
                  Jeff Donahue and
                  Karen Simonyan},
  title        = {Large Scale {GAN} Training for High Fidelity Natural Image Synthesis},
  booktitle    = {7th International Conference on Learning Representations, {ICLR} 2019,
                  New Orleans, LA, USA, May 6-9, 2019},
  year         = {2019}
}

@INPROCEEDINGS{karras1,
  author={Karras, Tero and Laine, Samuli and Aila, Timo},
  booktitle={2019 IEEE/CVF Conference on Computer Vision and Pattern Recognition (CVPR)}, 
  title={A Style-Based Generator Architecture for Generative Adversarial Networks}, 
  year={2019},
  volume={},
  number={},
  pages={4396-4405},
  doi={10.1109/CVPR.2019.00453}}

@INPROCEEDINGS{karras2,
  author={Karras, Tero and Laine, Samuli and Aittala, Miika and Hellsten, Janne and Lehtinen, Jaakko and Aila, Timo},
  booktitle={2020 IEEE/CVF Conference on Computer Vision and Pattern Recognition (CVPR)}, 
  title={Analyzing and Improving the Image Quality of StyleGAN}, 
  year={2020},
  volume={},
  number={},
  pages={8107-8116},
  doi={10.1109/CVPR42600.2020.00813}}

@inproceedings{improved_techs,
 author = {Song, Yang and Ermon, Stefano},
 booktitle = {Advances in Neural Information Processing Systems},
 editor = {H. Larochelle and M. Ranzato and R. Hadsell and M.F. Balcan and H. Lin},
 pages = {12438--12448},
 publisher = {Curran Associates, Inc.},
 title = {Improved Techniques for Training Score-Based Generative Models},
 url = {},
 volume = {33},
 year = {2020}
}

@inproceedings{isola2017image,
  title={Image-to-image translation with conditional adversarial networks},
  author={Isola, Phillip and Zhu, Jun-Yan and Zhou, Tinghui and Efros, Alexei A},
  booktitle={Proceedings of the IEEE conference on computer vision and pattern recognition},
  pages={1125--1134},
  year={2017}
}

@article{deveshwar2023synthesizing,
  title={Synthesizing Complex-Valued Multicoil MRI Data from Magnitude-Only Images},
  author={Deveshwar, Nikhil and Rajagopal, Abhejit and Sahin, Sule and Shimron, Efrat and Larson, Peder EZ},
  journal={Bioengineering},
  volume={10},
  number={3},
  pages={358},
  year={2023},
  publisher={MDPI}
}

@inproceedings{varnet,
  title={End-to-end variational networks for accelerated MRI reconstruction},
  author={Sriram, Anuroop and Zbontar, Jure and Murrell, Tullie and Defazio, Aaron et al.},
  booktitle={Medical Image Computing and Computer Assisted Intervention--MICCAI 2020: 23rd International Conference, Lima, Peru, October 4--8, 2020, Proceedings, Part II 23},
  pages={64--73},
  year={2020},
  organization={Springer}
}

@article{knoll2020fastmri,
  title={fastMRI: A publicly available raw k-space and DICOM dataset of knee images for accelerated MR image reconstruction using machine learning},
  author={Knoll, Florian and Zbontar, Jure and Sriram, Anuroop and Muckley, et al.},
  journal={Radiology: Artificial Intelligence},
  volume={2},
  number={1},
  pages={e190007},
  year={2020},
  publisher={Radiological Society of North America}
}

@article{fid_explanation2,
  title={A multimodal comparison of latent denoising diffusion probabilistic models and generative adversarial networks for medical image synthesis},
  author={M{\"u}ller-Franzes, Gustav and Niehues, Jan Moritz and Khader, Firas and Arasteh, Soroosh Tayebi and Haarburger, Christoph and Kuhl, Christiane and Wang, Tianci and Han, Tianyu and Nolte, Teresa and Nebelung, Sven and others},
  journal={Scientific Reports},
  volume={13},
  number={1},
  pages={12098},
  year={2023},
  publisher={Nature Publishing Group UK London}
}

@article{brain_tumor_seg,
  title={Brain tumor segmentation using synthetic MR images-A comparison of GANs and diffusion models},
  author={Usman Akbar, Muhammad and Larsson, M{\aa}ns and Blystad, Ida and Eklund, Anders},
  journal={Scientific Data},
  volume={11},
  number={1},
  pages={259},
  year={2024},
  publisher={Nature Publishing Group UK London}
}

@inproceedings{brain_gen,
  title={Brain imaging generation with latent diffusion models},
  author={Pinaya, Walter HL and Tudosiu, Petru-Daniel and Dafflon, Jessica and Da Costa, et al.},
  booktitle={MICCAI Workshop on Deep Generative Models},
  pages={117--126},
  year={2022},
  organization={Springer}
}

@article{gao2023synthetic,
  title={Synthetic data accelerates the development of generalizable learning-based algorithms for X-ray image analysis},
  author={Gao, Cong and Killeen, Benjamin D and Hu, Yicheng and Grupp, Robert B and Taylor, Russell H and Armand, Mehran and Unberath, Mathias},
  journal={Nature Machine Intelligence},
  volume={5},
  number={3},
  pages={294--308},
  year={2023},
  publisher={Nature Publishing Group UK London}
}

@article{al2023usability,
  title={On the usability of synthetic data for improving the robustness of deep learning-based segmentation of cardiac magnetic resonance images},
  author={Al Khalil, Yasmina and Amirrajab, Sina and Lorenz, et al.},
  journal={Medical Image Analysis},
  volume={84},
  pages={102688},
  year={2023},
  publisher={Elsevier}
}

@article{dhariwal2021diffusion,
  title={Diffusion models beat gans on image synthesis},
  author={Dhariwal, Prafulla and Nichol, Alexander},
  journal={Advances in neural information processing systems},
  volume={34},
  pages={8780--8794},
  year={2021}
}

@article{ssim,
  title={Image quality assessment: from error visibility to structural similarity},
  author={Wang, Zhou and Bovik, Alan C and Sheikh, Hamid R and Simoncelli, Eero P},
  journal={IEEE transactions on image processing},
  volume={13},
  number={4},
  pages={600--612},
  year={2004},
  publisher={IEEE}
}

@article{fid,
  title={Gans trained by a two time-scale update rule converge to a local nash equilibrium},
  author={Heusel, Martin and Ramsauer, Hubert and Unterthiner, Thomas and Nessler, Bernhard and Hochreiter, Sepp},
  journal={Advances in neural information processing systems},
  volume={30},
  year={2017}
}

@misc{fid_implement,
  author={Maximilian Seitzer},
  title={{pytorch-fid: FID Score for PyTorch}},
  month={August},
  year={2020},
  note={Version 0.3.0},
  howpublished={\url{https://github.com/mseitzer/pytorch-fid}},
}

@article{wang2023one,
  title={One for multiple: Physics-informed synthetic data boosts generalizable deep learning for fast MRI reconstruction},
  author={Wang, Zi and Yu, Xiaotong and Wang, Chengyan and Chen, Weibo and Wang, Jiazheng and Chu, Ying-Hua and Sun, Hongwei and others},
  journal={arXiv preprint arXiv:2307.13220},
  year={2023}
}

@article{goodfellow2020generative,
  title={Generative adversarial networks},
  author={Goodfellow, Ian and Pouget-Abadie, Jean and Mirza, Mehdi and Xu, Bing and Warde-Farley, David and Ozair, Sherjil and Courville, Aaron and Bengio, Yoshua},
  journal={Communications of the ACM},
  volume={63},
  number={11},
  pages={139--144},
  year={2020},
  publisher={ACM New York, NY, USA}
}

@inproceedings{bau2019seeing,
  title={Seeing what a gan cannot generate},
  author={Bau, David and Zhu, Jun-Yan and Wulff, Jonas and Peebles, William and Strobelt, Hendrik and Zhou, Bolei and Torralba, Antonio},
  booktitle={Proceedings of the IEEE/CVF International Conference on Computer Vision},
  pages={4502--4511},
  year={2019}
}

@inproceedings{maluleke2022studying,
  title={Studying bias in gans through the lens of race},
  author={Maluleke, Vongani H and Thakkar, Neerja and Brooks, Tim and Weber, et al.},
  booktitle={European Conference on Computer Vision},
  pages={344--360},
  year={2022},
  organization={Springer}
}

@article{peng2022generating,
  title={Generating Realistic 3D Brain MRIs Using a Conditional Diffusion Probabilistic Model},
  author={Peng, Wei and Adeli, Ehsan and Zhao, Qingyu and Pohl, Kilian M},
  journal={arXiv preprint arXiv:2212.08034},
  year={2022}
}

@inproceedings{jiang2023cola,
  title={CoLa-Diff: Conditional latent diffusion model for multi-modal MRI synthesis},
  author={Jiang, Lan and Mao, Ye and Wang, Xiangfeng and Chen, Xi and Li, Chao},
  booktitle={International Conference on Medical Image Computing and Computer-Assisted Intervention},
  pages={398--408},
  year={2023},
  organization={Springer}
}

@inproceedings{rombach2022high,
  title={High-resolution image synthesis with latent diffusion models},
  author={Rombach, Robin and Blattmann, Andreas and Lorenz, Dominik and Esser, Patrick and Ommer, Bj{\"o}rn},
  booktitle={Proceedings of the IEEE/CVF conference on computer vision and pattern recognition},
  pages={10684--10695},
  year={2022}
}

@inproceedings{adam,
  author       = {Diederik P. Kingma and
                  Jimmy Ba},
  editor       = {Yoshua Bengio and
                  Yann LeCun},
  title        = {Adam: {A} Method for Stochastic Optimization},
  booktitle    = {3rd International Conference on Learning Representations, {ICLR} 2015,
                  San Diego, CA, USA, May 7-9, 2015, Conference Track Proceedings},
  year         = {2015}
}
